\documentclass[letterpaper, 10 pt, conference]{ieeeconf}  

\IEEEoverridecommandlockouts                              

\usepackage{graphics} 
\usepackage{epsfig} 
\usepackage{mathptmx} 
\usepackage{times} 
\usepackage{amsmath} 
\usepackage{amssymb}  

\usepackage{graphicx}
\usepackage{multirow}
\usepackage[table,xcdraw]{xcolor}
\usepackage[normalem]{ulem}
\usepackage{array}
\usepackage{booktabs}
\usepackage{amsmath}
\usepackage{amsfonts}
\usepackage[backref]{hyperref}
\usepackage[compress]{cite}

\useunder{\uline}{\ul}{}

\def\eg{\emph{e.g}.}

\def\ie{\emph{i.e}.}

\def\para#1{\noindent{\bf{#1}}}

\usepackage{etoolbox}
\makeatletter
\patchcmd{\@makecaption}
  {\scshape}
  {}
  {}
  {}
\makeatletter
\patchcmd{\@makecaption}
  {\\}
  {.\ }
  {}
  {}
\makeatother

\begin{document}

\title{\LARGE \bf
Self-Supervised Ego-Motion Estimation Based on \\
Multi-Layer Fusion of RGB and Inferred Depth
}

\author{Zijie Jiang$^{1}$, Hajime Taira$^{1}$, Naoyuki Miyashita$^{2}$ and Masatoshi Okutomi$^{1}$
\thanks{$^{1}$Dept. of Systems and Control Engineering, Tokyo Institute of Technology,
        {\tt\small \{zjiang, htaira, mxo\}@ok.sc.e.titech.ac.jp
}}%
\thanks{$^{2}$R\&D Group, Olympus,
        {\tt\small naoyuki.miyashita@olympus.com
}}
}

\maketitle
\thispagestyle{empty}
\pagestyle{empty}

\begin{abstract}
In existing self-supervised depth and ego-motion estimation methods, ego-motion estimation is usually limited to only leveraging RGB information.
Recently, several methods have been proposed to further improve the accuracy of self-supervised ego-motion estimation by fusing information from other modalities, e.g., depth, acceleration, and angular velocity.
However, they rarely focus on how different fusion strategies affect performance.
In this paper, we investigate the effect of different fusion strategies for ego-motion estimation and propose a new framework for self-supervised learning of depth and ego-motion estimation, which performs ego-motion estimation by leveraging RGB and inferred depth information in a Multi-Layer Fusion manner.
As a result, we have achieved state-of-the-art performance among learning-based methods on the KITTI odometry benchmark.
Detailed studies on the design choices of leveraging inferred depth information and fusion strategies have also been carried out, which clearly demonstrate the advantages of our proposed framework\footnotemark[3]\footnotetext[3]{Our code will be available at \href{https://github.com/Beniko95J/MLF-VO}{https://github.com/Beniko95J/MLF-VO}.}.

\end{abstract}

\section{INTRODUCTION \label{sec:intro}}
\noindent
Structure from Motion (SfM) and Simultaneous Localization and Mapping (SLAM) are popular and promising techniques in computer vision. 
One key component in them is to get an accurate ego-motion between two consecutive frames, which is often carried as camera pose estimation using RGB images.
On top of the great success of classical methods based on 3D geometry and camera models, learning-based pose estimation methods~\cite{ummenhofer2017demon, zhou2017unsupervised} have recently got increasing research interests for their good fits to training data and feasibility in typical severe situations, \eg, poor lighting~\cite{li2021self}. 
Most of these works treat the pose estimation problem as a regression from input color images, and design pose estimators based on the convolutional neural network (CNN)~\cite{ummenhofer2017demon,ding2019camnet, zhou2017unsupervised,bian2019unsupervised,godard2019digging,li2019pose,ambrus2020two} or the recurrent neural network (RNN)~\cite{wang2017deepvo,wang2019recurrent,xue2019beyond,zou2020learning}. 

%
\begin{figure}[t]
    \centering
    {\small
    \begin{tabular}{c}
        \includegraphics[width=0.95\linewidth]{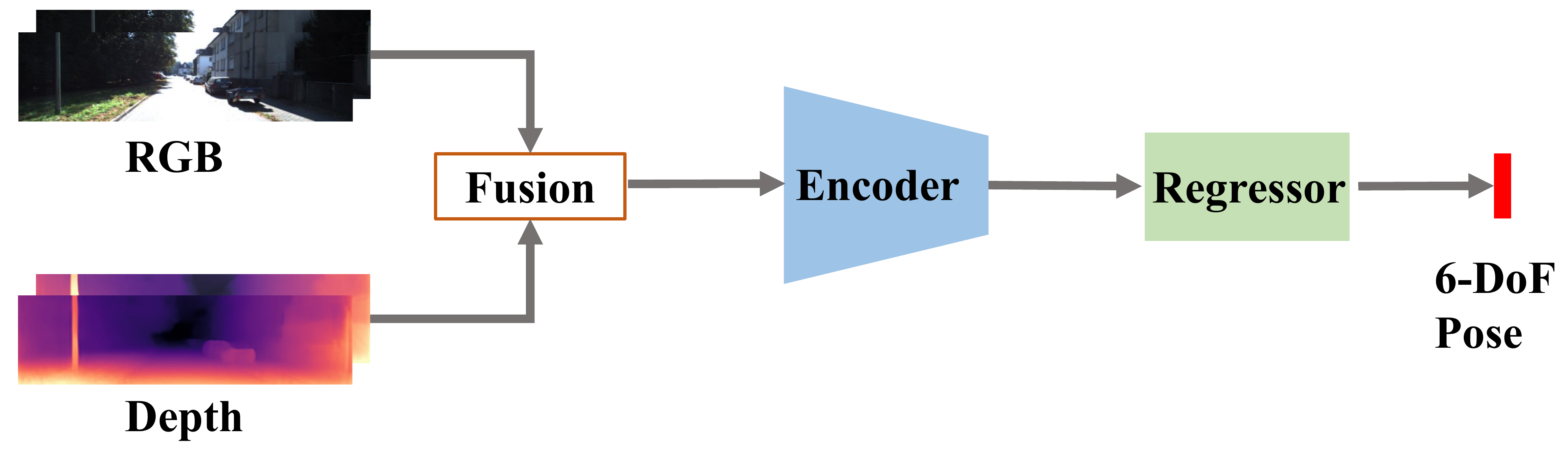} \\[-15pt]
        (a) Early fusion \\[13pt]
        \includegraphics[width=0.95\linewidth]{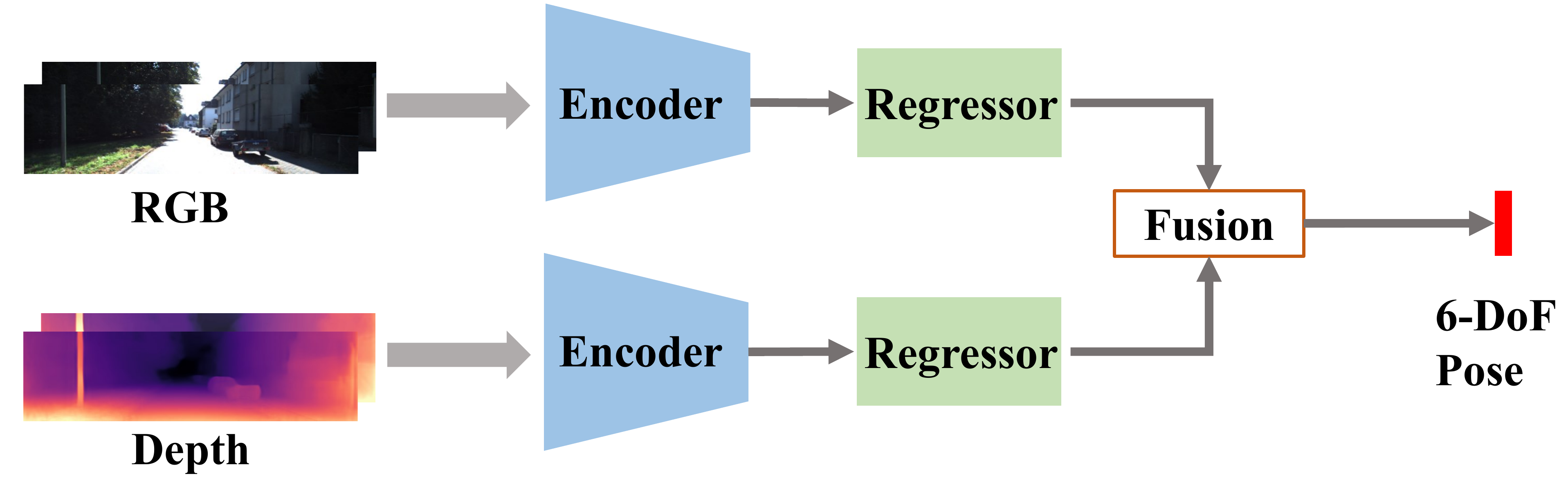} \\[-7pt]
        (b) Late fusion \\[10pt]
        \includegraphics[width=0.95\linewidth]{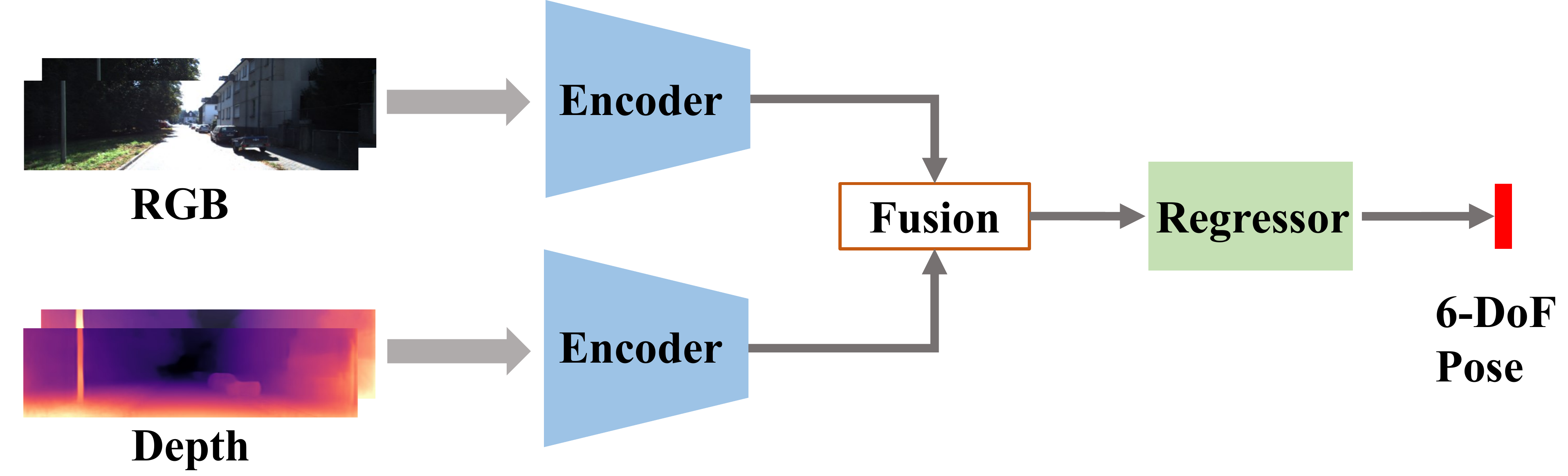} \\[-7pt]
        (c) Middle fusion \\[10pt]
        \includegraphics[width=0.95\linewidth]{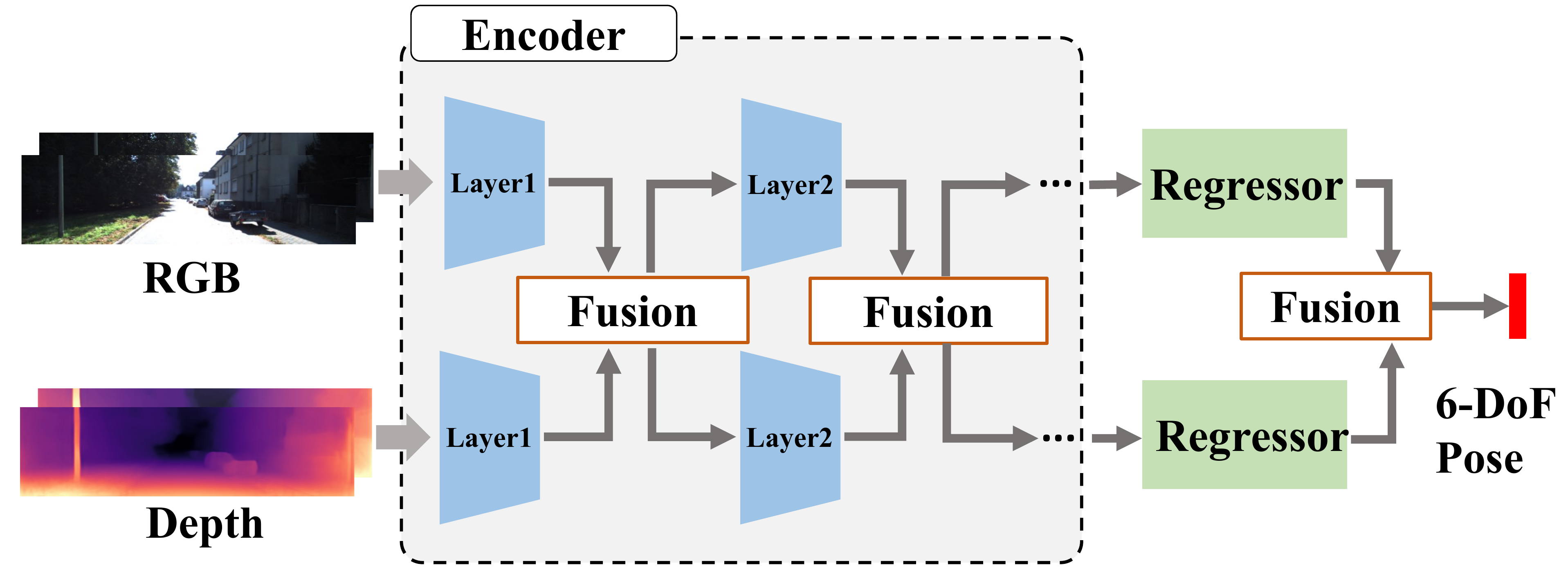} \\
        (d) Proposed multi-layer fusion (MLF-VO) \\[-5pt]
    \end{tabular}
    }
    \caption{Different fusion strategies depending on where to fuse RGB and depth modalities for pose estimation.}
    \label{fig_fusion_variation}
\end{figure}
One common challenge of those learned pose estimators is the capability to generalize to different situations, \eg, long-term scene changes including lighting changes~\cite{sattler2018benchmarking}. 
Several works~\cite{chen2019selective, han2019deepvio, wei2020unsupervised, li2021self} tackle this problem by adopting different modalities from additional sensors, such as LiDARs~\cite{li2021self} and IMUs~\cite{chen2019selective, han2019deepvio, wei2020unsupervised}, along with RGB images. 
Alternatively, self-supervised training strategies for the pose estimator have recently been studied~\cite{li2018undeepvo, zhan2018unsupervised, almalioglu2019ganvo} to utilize much more training data, which could consequently lead to a more generalized model. 
SfMLearner~\cite{zhou2017unsupervised} simultaneously learns to estimate ego-motion and scene structure, \ie, depth image, while measuring their consistency as a supervision. 
Furthermore, recent works~\cite{wang2019recurrent, ambrus2020two, zou2020learning, li2020self} reported the fact that feeding a pose estimator not only with original images but also with depth images predicted from RGB images can improve the performance, as well as fusing different modalities from real sensors. 

However, this approach raises one natural question: How can we effectively combine features from original color images and predicted {\it ``pseudo''} depth images? 
Since the prediction via CNN involves multiple levels of features in intermediate layers, there are various design choices (cf. Fig.~\ref{fig_fusion_variation}) to combine features from two streams, \eg, simply feed a pose estimator with concatenated RGB-D images, or combine features after encoding two modalities independently? 
To the best of our knowledge, there are few works that conduct thorough study to answer this question for the pose estimation task. 

In this paper, {\bf (1)} we design a baseline ego-motion estimation pipeline for two paired images, which can learn in a self-supervised manner. 
The pipeline firstly predicts a depth image for each source image, then feed all color and depth images to another pose estimator that predicts a relative pose between two cameras. 
{\bf (2)} Second, we conduct a study on this pipeline for comparing several design choices for fusing two modalities.
{\bf (3)} Based on the study, we propose an RGB-D-to-pose estimator that fuses two modalities in an incremental fashion.
As shown in Fig.~\ref{fig_fusion_variation}~(d), our pose estimator, MLF-VO (Visual Odometry using Multi-Layer Fusion), namely, consists of two streams for color and depth inputs and several fusion layers that combine intermediate features at multiple levels. 
Whereas the core architecture of the fusion layer is originated from an existing work~\cite{wang2020deep}, we also propose a new regularization loss to effectively learn the model.
{\bf (4)} Finally, our overall pipeline achieves better pose accuracy than existing works on KITTI Odometry benchmark~\cite{geiger2012we,geiger2013vision}, while requiring less computational time.

\section{Related Work \label{sec:relatedworks}}

\subsection{Self-supervised learning of depth and ego-motion}

\noindent
Self-supervised learning of depth and ego-motion from monocular video is originally offered in \cite{zhou2017unsupervised}, which proposes to utilize two decoupled networks to estimate depth and ego-motion independently, and get supervisory signals by minimizing the photometric loss between the synthetic and original image~\cite{jaderberg2015spatial}. 
Building upon this paradigm, recent works focusing on the improvement of self-supervised depth estimation \cite{godard2019digging} have achieved exciting progress, showing competitive performance compared to supervised methods.

On the other hand, there are relatively few works focusing on the improvement of ego-motion estimation.
\cite{nabavi2020unsupervised} extends the standard ego-motion estimation network to incorporate feedback through iterative view synthesis.
Instead of direct regression model for ego-motion, \cite{zhao2020towards} proposes to estimate an optical flow between two images and solves ego-motion as an optimal fundamental matrix. 
Besides these works which still perform ego-motion estimation purely in the RGB domain, some other works attempt to introduce data from other sensors, \eg, LiDAR \cite{li2021self} and IMU \cite{chen2019selective, wei2020unsupervised}, as additional inputs to improve the ego-motion accuracy in the spirit of sensor fusion. 
Following the idea of performing motion estimation in a mixed domain, \cite{ambrus2020two} further proposes a two-stream network that leverages the original RGB images and the internally inferred depth maps as inputs of ego-motion network. 
Our method shares the idea with \cite{ambrus2020two}, but we further investigate how different fusion strategies affect the final performance and propose a new relative pose estimation network based on multi-layer fusion of RGB and inferred depth information. 

\subsection{Multi-modal fusion}
\noindent
Early studies of multi-modal fusion~\cite{snoek2005early, atrey2010multimodal,bruni2014multimodal} categorizes the fusion strategy into two broad types: early (raw-level) fusion and late (decision-level) fusion. 
Recent deep learning literature has also been studied mainly in either of these two branches~\cite{lazaridou2014wampimuk,ramachandram2017deep,baltruvsaitis2018multimodal}. 
Early fusion, which aggregates multiple modalities before making decision, is often performed as a concatenation along the input channels~\cite{zhang2018deep, zeng2019deep} or averaging~\cite{hazirbas2016fusenet}, where the final decision is made by the subsequent single encoder and regressor. 
On the other hand, late fusion strategy makes decisions from each modality solely. 
The final decision is obtained as an ensemble of multiple outputs, which can be carried out by averaging~\cite{ambrus2020two} or a learned meta-model~\cite{glodek2011multiple,wang2020deep}. 
As an intersection of these two approaches, middle (feature-level) fusion has also been developed~\cite{ramachandram2017deep,chen2019selective,wei2020unsupervised,li2021self}. 
This approach prepares a CNN or RNN-based encoder for each modality to obtain deeply encoded features. 
Features are then combined via a subsequent fusion layer, such as self-attention mechanism~\cite{hori2017attention,vaswani2017attention}, and fed to the final regressor. 
%
We investigate all three fusion strategies in our ego-motion estimation pipeline and propose a new multi-layer fusion strategy. 

%
%
\begin{figure*}[t]
    \centering
    \begin{tabular}{c}
        \includegraphics[width=0.95\linewidth]{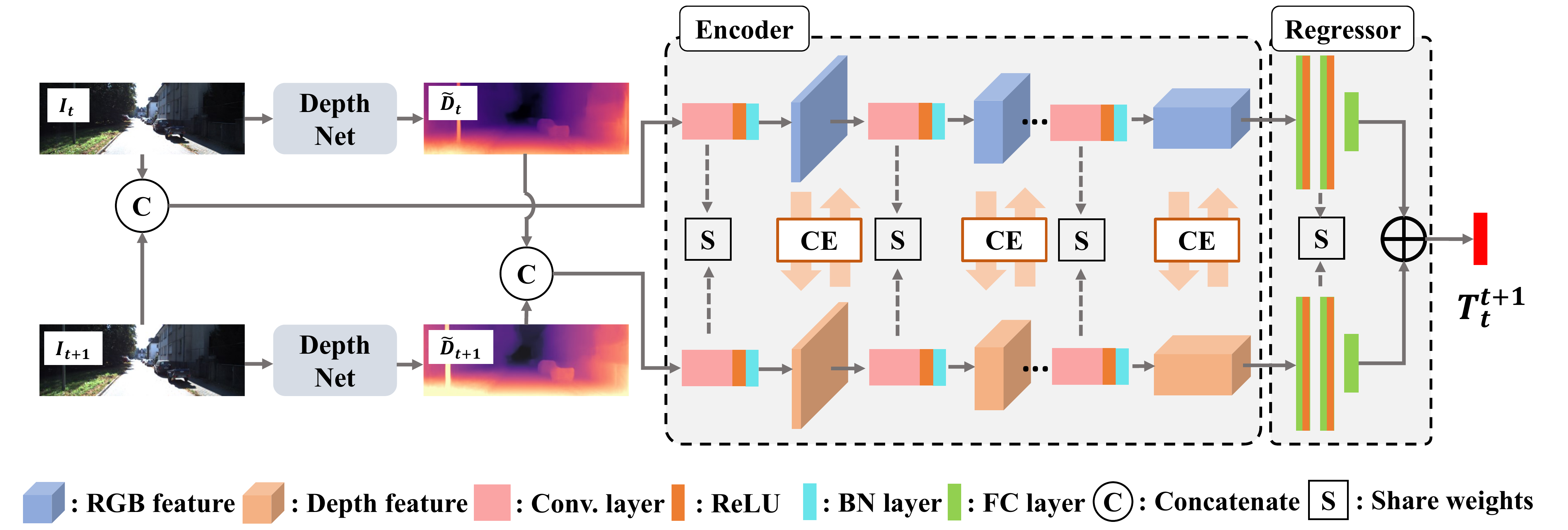} \\[-3pt]
        (a) Forward pass of proposed network (MLF-VO) \\[3pt]
        \includegraphics[width=0.95\linewidth]{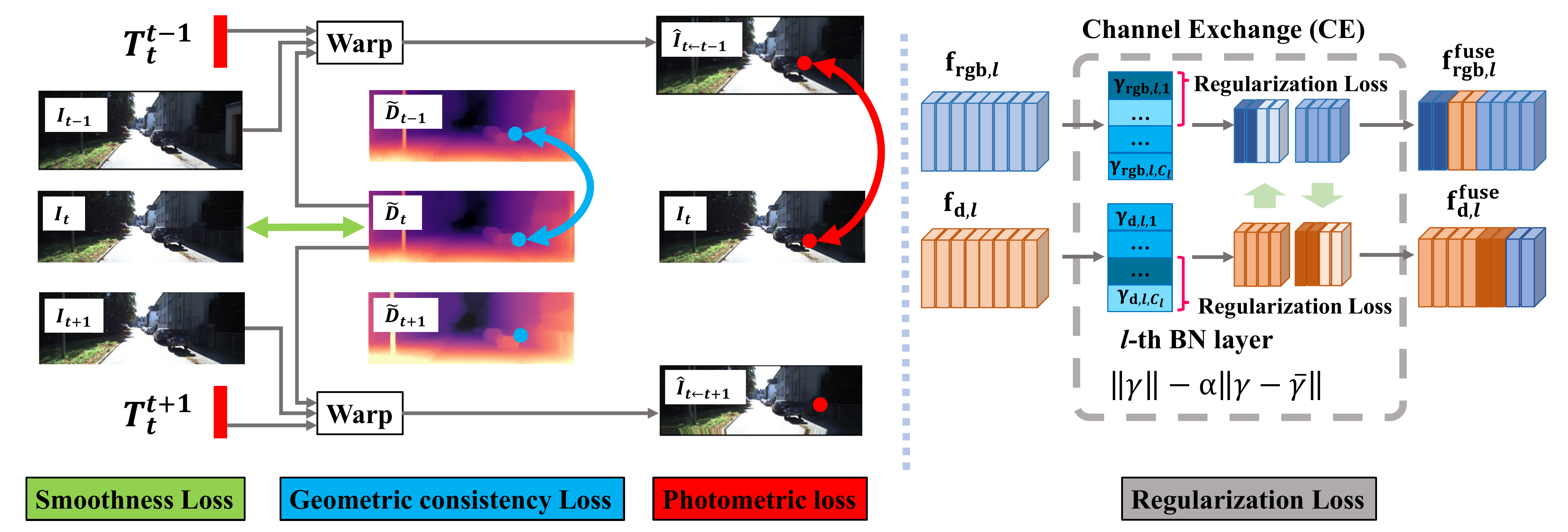} \\[-3pt]
        (b) Self-supervised training loss 
    \end{tabular}
    \caption{{\bf Overview}. (a) Our MLF-VO consists of two streams of CNN encoder and regressor fed with RGB and inferred depth images. After each BN layer of the encoder, intermediate features from two modalities are fused by
    CE operation. (b) Self-supervised training for our MLF-VO is based on the consistency of warped RGB (Photometric loss) and depth images (Geometric consistency loss). We additionally introduce a regularization loss that controls scaling factors of BN layers which are used in CE operation. }
    \label{fig_overview}
\end{figure*}
\section{Proposed Method \label{sec_method}}
\noindent
Fig.~\ref{fig_overview} illustrates the overview of our proposed ego-motion estimation framework (MLF-VO) and the training manner for the framework. 
In this section, we firstly introduce our baseline framework for ego-motion estimation composed of two independent CNN models for depth prediction and relative pose estimation (Sec.~\ref{subsec_network}). 
Next, we detail our proposed relative pose estimator that fuses encoded features of color and inferred depth images based on the channel exchange in multiple stages of the encoder (Sec.~\ref{subsec_estimator}). 
Finally, Sec.~\ref{subsec_loss} presents our self-supervised joint learning procedure for depth and relative pose estimators, along with a new regularization loss that encourages the cross-modal information exchanges. 

\subsection{Baseline ego-motion estimation pipeline \label{subsec_network}}
\noindent
Fig.~\ref{fig_overview}~(a) illustrates our ego-motion estimation framework. 
Our framework consists of two independent prediction models: $\theta_{{\rm depth}}$ for depth prediction and $\theta_{{\rm pose}}$ for relative pose estimation. 
Assuming an image pair of consecutive video frames $\left\{I_{t}, I_{t+1}\right\}$, our pipeline firstly estimates depth images along with each input frame: 
\begin{equation}
    D_{t} = \theta_{{\rm depth}}(I_t), \; D_{t+1} = \theta_{{\rm depth}}(I_{t+1}),
\end{equation}
where $D_t$ is the pixel-aligned inverse depth map for $I_{t}$. 
We construct the depth prediction model in the same manner as in \cite{godard2019digging}, which is based on the fully convolutional U-Net architecture obtaining depths at four scales.

Next, the subsequent relative pose estimator $\theta_{{\rm pose}}$ predicts a relative pose between consecutive frames as a final output of ego-motion: 
\begin{equation}
    T^{t+1}_{t} = \theta_{{\rm pose}}(I_t\oplus I_{t+1}, D_{t}\oplus D_{t+1}).
\end{equation}
Here, $\oplus$ denotes a concatenation along input channels.
In contrast to several existing works regressing an ego-motion only from the original color image~\cite{zhou2017unsupervised, bian2019unsupervised, godard2019digging}, we feed the model also with the inferred depth maps, which can provide additional information that is useful in some typical situations (cf. Fig.~\ref{fig_qual}). 
Therefore, the fusion strategy for color and depth modalities plays an important role to estimate an accurate ego-motion. 


\subsection{Relative pose estimation based on multi-layer fusion \label{subsec_estimator}}
\noindent
As discussed in Sec.~\ref{sec:intro} and Sec.~\ref{sec:relatedworks}, there are various design choices to combine two different modalities into the final relative pose output. 
Based on a study of possible strategies applied in our baseline framework (Sec.~\ref{subsec_study}), we finally decide to employ a {\it multi-layer fusion strategy} that fuses several features appearing in intermediate layers of the encoder (Fig.~\ref{fig_overview}~(a)). 

Our pose estimation model has a two-stream structure for encoding features from each of RGB and inferred depth inputs.
To effectively fuse multiple levels of features while remaining complementary characteristics of RGB and depth modalities, we exploit the Channel Exchange (CE) strategy~\cite{wang2020deep} that swaps feature elements based on its importance. 
We employ the ResNet-18 architecture~\cite{he2016deep} as encoders for both streams, while sharing all weights except for Batch-Normalization (BN) layers so that we can ensure both features before BN are in the same latent space, \ie, all feature elements can be replaced with the ones from the other modality. 
In every BN layer, CE evaluates the importance of each top (for color) or bottom (for depth) half of feature channels as the BN scaling factor $\gamma$. 
We assume the channels with smaller $\gamma$ than the threshold $0.02$ become redundant to the final outputs and thus replace them with the same channels extracted from the other modality. 
After encoding each modality into the common feature space, 
we regress each feature into a 6-dimensional relative pose representation via a pose regressor composed of two fully-connected layers with ReLU activation and a final fully-connected layer. 
The final output of relative pose is obtained as a weighted sum of the outputs of each stream, where the weighting factor is jointly learned during the training. 

Whereas the CE strategy originally has been tested on sensor fusion~\cite{wang2020deep}, we employ it to the fusion of color and inferred depth map. 
We also propose a new training loss for CE that prevents the model from reaching a singular solution, which will be introduced in the next subsection.

\subsection{Self-supervised training \label{subsec_loss}}
\noindent
\para{Self-supervised loss for depth and motion prediction}. We train our MLF-VO in a self-supervised manner similar to \cite{zhou2017unsupervised,godard2019digging}, which constructs a supervisory signal as the consistency of predicted motion and scene structure (Fig.~\ref{fig_overview}~(b)). 
Assuming an input of target and source images $\{I_t,I_s\}$, the depth map $D_t$ and the relative motion $T_{t}^{s}$, which are obtained from a forward pass, can present pixel-wise correspondences between two images, \ie, we can generate a synthetic target view $\hat{I}_{t\leftarrow s}$ by projecting source image pixels onto the target image plane~\cite{jaderberg2015spatial}. 
Therefore, the confidence of predictions can be evaluated by the photometric loss $L_p$ that evaluates a photometric error~\cite{godard2019digging} of the synthetic target view. 
Similarly, \cite{bian2019unsupervised} introduces the geometric consistency loss $L_{gc}$ that
evaluates the consistency of predicted depth maps by projecting scene points defined from a depth map instead of image pixels. 
We also adopt a smoothness loss for the depth map $L_s(I_t,D_t)$ in the same manner as in \cite{godard2017unsupervised}. 

\para{Regularization loss for CE}. \cite{wang2020deep} originally imposes $l_1$ norm penalization on BN scaling factors to learn the redundant channels in CE. 
Although the $l_1$ norm loss is simple yet effective, there are still certain possibilities that the solution falls into a singular point, \ie, all scaling factors converge to zeros thus all possible channels are exchanged. 
Therefore, we extend the original $l_1$ norm with the polarization regularizer proposed in \cite{zhuang2020neuron} as:
\begin{align}
L_r=\sum_m\sum_l\sum_i \Vert \gamma_{m,l,i} \Vert - \alpha\Vert \gamma_{m,l,i}-\hat{\gamma}_{l} \Vert
\end{align}
where $\hat{\gamma}_{l}$ denotes the mean of exchangeable scaling factors in the $l$-th BN layer. 
$\gamma_{m,l,i}$ denotes the scaling factor of the $i$-th exchangeable channel of the $l$-th BN layer of modality $m\in \left\{rgb, depth\right\}$.

\para{End-to-end training}. Motivated by per-pixel minimum loss strategy~\cite{godard2019digging}, which effectively tackles the problem of occluded regions in training, we prepare a set of training snippets $\left\{I_t, I_s, s\in\left\{t-1, t+1\right\}\right\}$ composed of one target image $I_t$ and its consecutive source images $\{I_s\}$. 
The photometric loss $\bar{L_p}$ and geometric consistency loss $\bar{L_{gc}}$ are then re-formulated: 
\begin{equation}
\begin{aligned}
\bar{L_p}(I_t,I_s,T_{t}^{s})=\mathop{{\rm min}}\limits_{s}L_p(I_t,\hat{I}_{t\leftarrow s}), \\
\bar{L_{gc}}(D_t,D_s,T_{t}^{s})=\mathop{{\rm min}}\limits_{s}L_{gc}(D_t,D_s,T_{t}^{s}).
\end{aligned}
\end{equation}
Also exploiting the multi-scale predictions of the depth map, $\bar{L_p}$, $\bar{L_{gc}}$ and $L_s$ are minimized over 4 output scales. 
Finally, the total self-supervised loss is formulated as:
\begin{align}
    \begin{aligned}
    L_{self}=\bar{L_p}(I_t,&I_s,T^{s}_{t})+\lambda_1\bar{L_{gc}}(D_t,D_s,T^{s}_{t})+ \\
    &\lambda_2L_s(I_t,D_t)+\lambda_3L_r.
    \end{aligned}
\end{align}
Using this joint loss, we simultaneously learn both depth prediction and relative pose estimation model. We will provide details of our training setting on KITTI Odometry dataset in Sec.~\ref{sec:exp}. 
\section{Experiments \label{sec:exp}}
\noindent
%
\subsection{Experimental Setup \label{sec_setup}}
\para{Datasets}: We adopt the widely used KITTI Odometry benchmark \cite{geiger2012we, geiger2013vision} to evaluate our method.
The benchmark contains 22 urban and highway driving sequences, among which only 11 sequences (Sequence 00-10) have ground-truth trajectory labels from GPS/IMU readings. 
Following \cite{bian2019unsupervised, ambrus2020two, zou2020learning}, we select Sequence 00-08 as training data, and test the trained model on Sequence 09 and 10.
In addition, we also report the results on the rest sequences of KITTI Odometry benchmark in the same manner as in \cite{zou2020learning}, which run the stereo version of ORB-SLAM2 to obtain reference trajectories.

\para{Evaluation Metrics}:
We adopt the KITTI Odometry criterion, which reports the average translational error $T_{rel}(\%)$ and rotational errors $R_{rel}(^\circ/100m)$ of possible sub-sequences of length (100, 200, $\cdots$, 800) meters, as the main evaluation criteria.
We also report the translational RMSE of the whole trajectory to evaluate the global trajectory accuracy.
For self-supervised monocular methods, since the absolute scale is unknown, we scale and align the predicted trajectory to the ground-truth associated poses using \cite{umeyama1991least} before evaluation.

\para{Implementation}: We implement our system based on the PyTorch framework \cite{paszke2017automatic}.
Both the depth and pose networks receive input images of size $640\times 192$ pixels.
The batch size is set to 12 and the model is trained for 40 epochs. 
The learning rate is set to 1e-4 for the first 20 epochs and drops to 5e-5 for the remaining epochs.
We train our full model using a single NVIDIA RTX 3090 and test the model using a single NVIDIA GTX 1080Ti.
We empirically set the hyper-parameters as follows: $\lambda_1$=1e-2, $\lambda_2$=1e-3, $\lambda_3$=2e-5, $\alpha$=1e-1.

\subsection{Comparison of different fusion strategies \label{subsec_study}}
\begin{table*}[t]
\caption{Comparison among different variants on sequence 09 and 10 of the KITTI Odometry dataset \cite{geiger2012we}. The best performance is in \textbf{bold}.}
\renewcommand\arraystretch{1.2}
\centering
{\normalsize
\begin{tabular}{ll|ccc|ccc|ccc}
\hline
                           &                          & \multicolumn{3}{c|}{Seq. 09}                                                                                       & \multicolumn{3}{c|}{Seq. 10}                                                                                       & \multicolumn{3}{c}{Avg.}                                                                                          \\ \cline{3-11} 
 &  & RMSE & $T_{rel}$ & $R_{rel}$ & RMSE & $T_{rel}$ & $R_{rel}$ & RMSE & $T_{rel}$ & $R_{rel}$ \\[-2pt]
Modality & Fusion & (m) & (\%) & (deg/100m) & (m) & (\%) & (deg/100m) & (m) & (\%) & (deg/100m) \\ \hline
RGB                        & -                        & 11.10                                & 4.97                                 & 1.72                                 & 11.58                                & 6.45                                 & 2.52                                 & 11.34                                & 5.71                                 & 2.12                                 \\
Depth                      & -                        & 13.54                                & 4.66                                 & 1.93                                 & {\color[HTML]{000000} \textbf{6.04}}                                 & 4.94                                 & 1.75                                 & 9.79                                & 4.80                                 & 1.84                                 \\
RGB+Depth                  & Early                    & 13.77                                & 5.22                                 & 1.79                                 & 10.42                                & 5.56                                 & 2.13                                 & 12.10                                & 5.39                                 & 1.96                                 \\
RGB+Depth                  & Late                     & 11.25                                & 4.91                                 & {\color[HTML]{000000} 1.79}          & 7.91                                 & 5.67                                 & 1.71                                 & 9.58                                 & 5.29                                 & 1.75                                 \\
RGB+Depth                  & Middle                     & 10.37                                & 4.32                                 & {\color[HTML]{000000} \textbf{1.37}}          & 8.06                                 & 5.14                                 & 1.62                                 & 9.21                                 & 4.73                                 & 1.50                                 \\
RGB+Depth                  & Multi-layer                  & {\color[HTML]{000000} \textbf{9.86}} & {\color[HTML]{000000} \textbf{3.90}} & {\color[HTML]{000000} 1.41} & 7.36 & {\color[HTML]{000000} \textbf{4.88}} & {\color[HTML]{000000} \textbf{1.38}} & {\color[HTML]{000000} \textbf{8.61}} & {\color[HTML]{000000} \textbf{4.39}} & {\color[HTML]{000000} \textbf{1.39}} \\ \hline
\end{tabular}
}
\label{table_variants}
\end{table*}
%
\noindent
As mentioned in Sec.~\ref{subsec_estimator}, we first evaluate various design choices of our relative pose estimators that leverage color and depth information.
According to the stage of fusions (cf. Fig.~\ref{fig_fusion_variation}), we build 5 variants of pose estimator, and compare them with our proposed multi-layer fusion model on Sequence 09 and 10 of the KITTI Odometry dataset (Tab.~\ref{table_variants}). 
The first two variants are: 1) RGB: the ego-motion network only takes RGB as inputs. 2) Depth: the ego-motion network only takes depth as inputs.
The rest variants take both RGB and depth as inputs, but leveraging different fusion strategies: 3) RGB+Depth (early fusion): the RGB and depth inputs are concatenated along channel dimension as an integrated input to a single ego-motion network.
4) RGB+Depth (late fusion): the RGB and depth streams are processed by two separate ego-motion networks and the outputs are combined using jointly learned weighting factors, which are normalized by an additional softmax layer~\cite{wang2020deep}.
5) RGB+Depth (middle fusion): the RGB and depth features encoded from two separate encoders are fused before a single pose regressor using Soft Fusion~\cite{chen2019selective}.
6) RGB+Depth (multi-layer fusion): our proposal as introduced in Sec. \ref{subsec_estimator}. 
For a justified comparison, we use the ResNet18 architecture as the encoder for all variants and the regressors are all implemented as two fully-connected layers with ReLU activation and a final fully-connected layer.
The first 5 variants are trained using the same loss function without regularization loss for CE.

One interesting observation from the validation results is that the early and late fusion methods sometimes show worse performance than the methods leveraging only single modality. 
This result implies the importance of selecting an appropriate fusion strategy. 
Middle fusion provides rather better results than the early and late fusion methods, while performing fusion for encoded features that include higher-semantic information than the final decision. 
Multi-layer fusion strategy gives the best performance among these comparisons. 
We attribute the results to the fact that the CE operation in multiple layers evaluates encoded features from both coarse (more semantic) and fine (more structural) layers, which further exploit the complementary properties between RGB and depth information. 
Based on these observations, we finally decide to employ the CE-based multi-layer fusion strategy for our relative pose estimation model.
In the following comparisons, we refer to the proposed multi-layer fusion model as "MLF-VO".

\begin{figure}[t]
    \centering
    \includegraphics[width=1\linewidth]{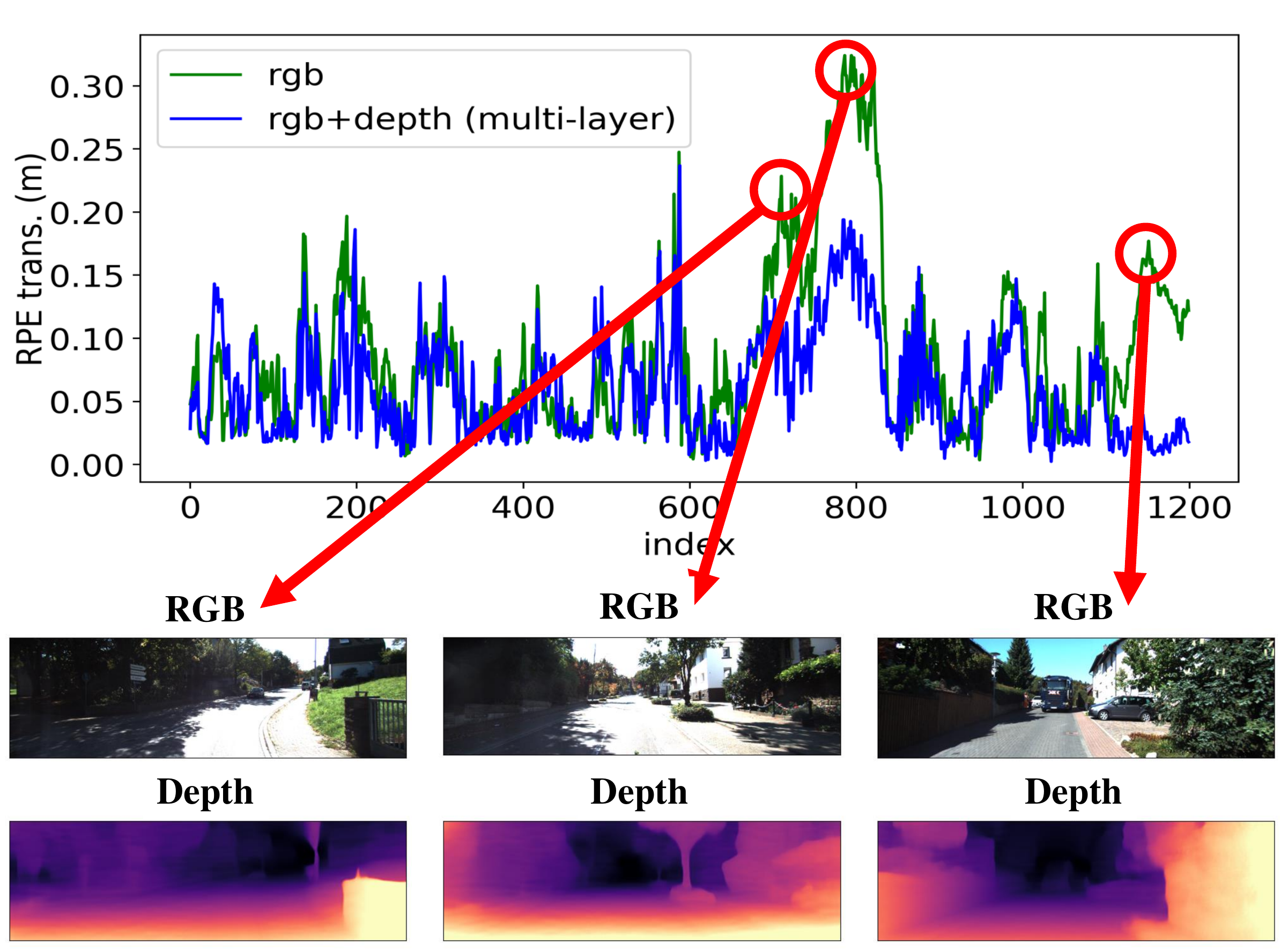}
    \caption{The top is the curve of relative translational errors of consecutive frames on Sequence 10 of KITTI Odometry benchmark. The bottom lists RGB images and inferred depth map where the our model performs explicitly better than the RGB-only model.}
    \label{fig_qual}
\end{figure}

Fig.~\ref{fig_qual} shows visual examples where fusing depth modality can help to obtain more accurate ego-motion. 
Ego-motion estimated using only RGB images results in relatively large pose errors in some specific sections. 
We attribute the failures to the lighting conditions of the sections, which produces a heavily saturated regions in the input RGB image. 
On the other hand, inferred depth map can still provides structural information of these scenes. 
Therefore, our ego-motion estimation utilizing both of RGB and depth images largely improves the accuracy of estimated motion. 
These results also demonstrate that our multi-layer fusion in relative pose estimator can retain complementary information extracted from different modalities for pose estimation. 

\subsection{Comparison with other methods}
\noindent
In Table.~\ref{table_others}, we compare our best model with several existing methods, including the monocular version of ORB-SLAM2~\cite{mur2017orb} which employs a classic geometry-based estimation (Geo.), supervised learning methods (Sup.)~\cite{wang2017deepvo, xue2019beyond} and self-supervised learning methods (Self Sup.)~\cite{li2019pose, godard2019digging, bian2019unsupervised, ambrus2020two, zou2020learning}. 
The results of ORB-SLAM2-M methods are the median of 5 runs. 
For Monodepth2~\cite{godard2019digging}, we use a pre-trained models provided by authors. For the remaining methods, we take the results reported in their paper. 
'-' means the results are not available from that paper. 
Note that all supervised methods are trained on Sequence 00, 02, 08, 09 of the KITTI Odometry dataset, while the self-supervised methods are trained using the same data split as ours.

Our MLF-VO achieves the best average performance among other self-supervised methods.
Compared to Monodepth2~\cite{godard2019digging} and SC-SfMLearner~\cite{bian2019unsupervised}, which are trained by a similar loss to ours, our final model outperforms them by a significant margin in the task of ego-motion estimation. 
TSN~\cite{ambrus2020two} consists of a two-stream network for RGB and depth as like in ours. Our method still shows superior results demonstrating that the fusion strategy plays an important role in the design of network utilizing two or more modalities as inputs.
LTMVO~\cite{zou2020learning} introduces LSTM structures to capture temporal information from color and depth images. 
Compared to this state-of-the-art ego-motion estimation pipeline, proposed MLF-VO also provides comparable accuracy of estimated motions, while requiring less computation (shown below).
Furthermore, proposed method is also comparable with the geometric method and supervised methods. 

{\tabcolsep=3pt
\begin{table*}[t]
\caption{Odometry results compared with the state-of-the-art methods. The best results of each block are highlighted by \textbf{bold} style. In the bottom block (Self Sup.), the best and the second best results are highlighted by {\bf {\color[HTML]{FE0000} red}} and {\color[HTML]{3531FF} blue} characters, respectively.}
\renewcommand\arraystretch{1.2}
\centering
\resizebox{\linewidth}{!}
{
\begin{tabular}{ll|ccc|ccc|ccc}
\hline
                                                 &                             & \multicolumn{3}{c|}{Seq. 09}                                                                                       & \multicolumn{3}{c|}{Seq.10}                                                                                        & \multicolumn{3}{c}{Avg.}                                                                                          \\ \cline{3-11} 
                                                 &  & RMSE & $T_{rel}$ & $R_{rel}$ & RMSE & $T_{rel}$ & $R_{rel}$ & RMSE & $T_{rel}$ & $R_{rel}$ \\[-2pt]
                                                 & \multicolumn{1}{c|}{Method} & (m) & (\%) & (deg/100m) & (m) & (\%) & (deg/100m) & (m) & (\%) & (deg/100m)\\[2pt] \hline
\multicolumn{1}{l|}{}                            & ORB-SLAM2-M (w/o LC) \cite{mur2017orb}       & {\color[HTML]{000000} 41.75}         & {\color[HTML]{000000} 10.03}          & {\color[HTML]{000000} \textbf{0.29}}          & 7.74 & 3.64 & {\color[HTML]{000000} \textbf{0.32}}          & 24.75                             & 6.84                                 & {\color[HTML]{000000} \textbf{0.31}}                                 \\
\multicolumn{1}{l|}{\multirow{-2}{*}{Geo.}}      & ORB-SLAM2-M (w LC) \cite{mur2017orb}                & {\color[HTML]{000000} \textbf{9.84}} & {\color[HTML]{000000} \textbf{3.48}} & 0.39 & {\color[HTML]{000000} \textbf{7.10}}          & {\color[HTML]{000000} \textbf{3.46}}          & 0.38 & {\color[HTML]{000000} \textbf{8.47}} & {\color[HTML]{000000} \textbf{3.47}} & 0.39 \\ \hline
\multicolumn{1}{l|}{}                            & DeepVO \cite{wang2017deepvo}                     & -                                    & -                                    & -                                    & -                                    & 8.11                                 & 8.83                                 & -                                    & -                                    & -                                    \\
\multicolumn{1}{l|}{\multirow{-2}{*}{Sup.}}      & BeyondTracking \cite{xue2019beyond}             & -                                    & -                                    & -                                    & -                                    & {\color[HTML]{000000} \textbf{3.94}} & {\color[HTML]{000000} \textbf{1.72}} & -                                    & -                                    & -                                    \\ \hline
\multicolumn{1}{l|}{}                            & PoseGraph \cite{li2019pose}                   & -                                    & 8.10                                 & 2.81                                 & -                                    & 12.90                                & 3.17                                 & -                                    & 10.50                                & 2.99                                 \\
\multicolumn{1}{l|}{}                            & Monodepth2 \cite{godard2019digging}                 & 76.42                                & 17.22                                & 3.86                                 & 20.47                                & 11.72                                & 5.35                                 & 48.45                                & 14.47                                & 4.61                                 \\
\multicolumn{1}{l|}{}                            & SC-SfMLearner \cite{bian2019unsupervised}               & -                                    & 11.20                                & 3.35                                 & -                                    & 10.10                                & 4.96                                 & -                                    & 10.65                                & 4.16                                 \\
\multicolumn{1}{l|}{}                            & TSN \cite{ambrus2020two}                        & -                                    & 6.72                                 & 1.69                                 & -                                    & 9.52                                 & 1.59                                 & -                                    & 8.12                                 & 1.64                                 \\
\multicolumn{1}{l|}{}                            & LTMVO \cite{zou2020learning}                       & {\color[HTML]{3531FF} 11.30}         & {\color[HTML]{FE0000} \textbf{3.49}} & {\color[HTML]{FE0000} \textbf{1.00}} & {\color[HTML]{3531FF} 11.80}         & {\color[HTML]{3531FF} 5.81}          & {\color[HTML]{3531FF} 1.80}          & {\color[HTML]{3531FF} 11.55}         & {\color[HTML]{3531FF} 4.65}          & {\color[HTML]{3531FF} 1.40}          \\
\multicolumn{1}{l|}{\multirow{-6}{*}{Self Sup.}} & {\bf MLF-VO (Ours)}                        & {\color[HTML]{FE0000} \textbf{9.86}} & {\color[HTML]{3531FF} 3.90}          & {\color[HTML]{3531FF} 1.41}          & {\color[HTML]{FE0000} \textbf{7.36}} & {\color[HTML]{FE0000} \textbf{4.88}} & {\color[HTML]{FE0000} \textbf{1.38}} & {\color[HTML]{FE0000} \textbf{8.61}} & {\color[HTML]{FE0000} \textbf{4.39}} & {\color[HTML]{FE0000} \textbf{1.39}} \\ \hline
\end{tabular}
}
\label{table_others}
\end{table*}
}

\begin{figure}[t]
    \centering
    \begin{tabular}{cc}
        \includegraphics[width=0.45\linewidth]{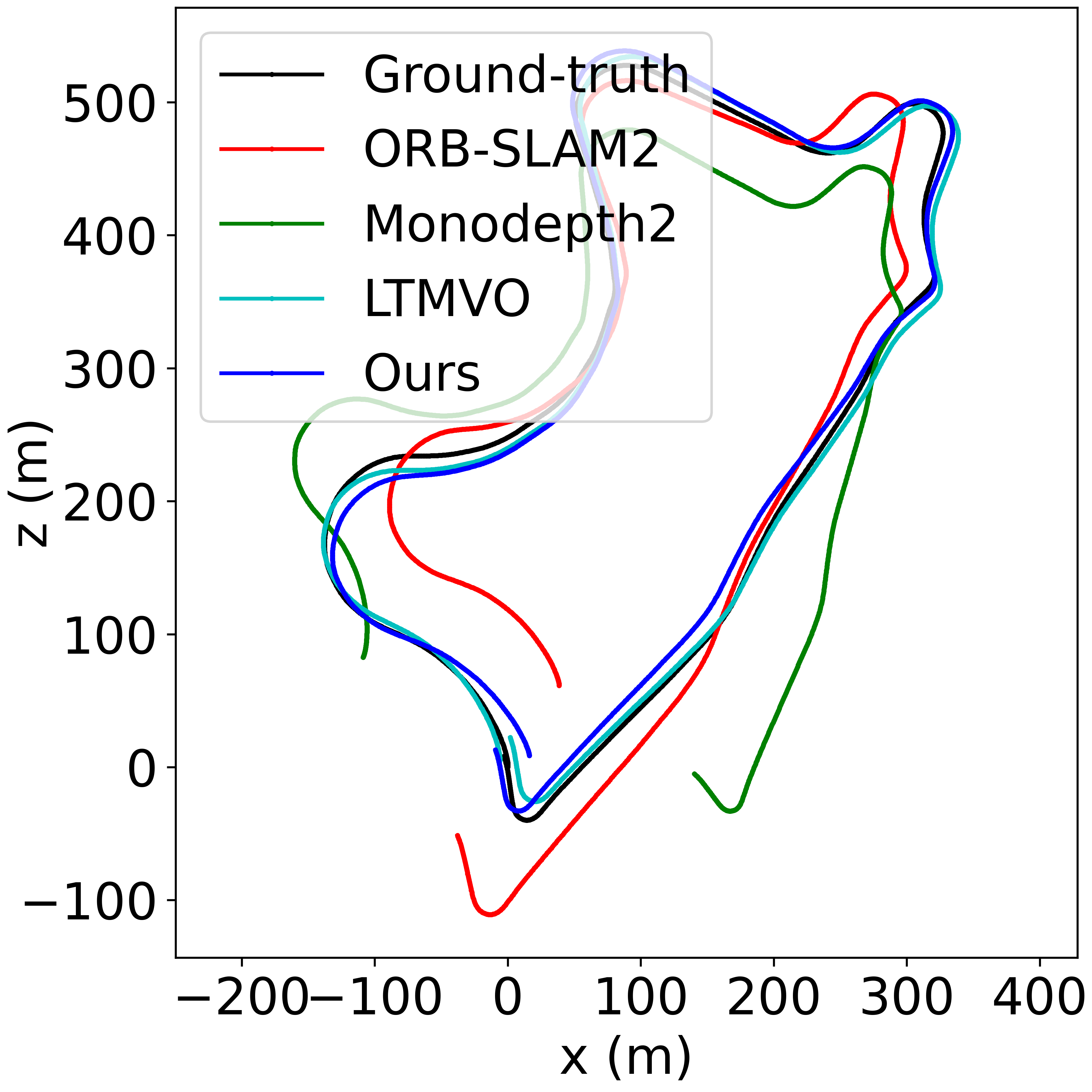} & 
        \includegraphics[width=0.45\linewidth]{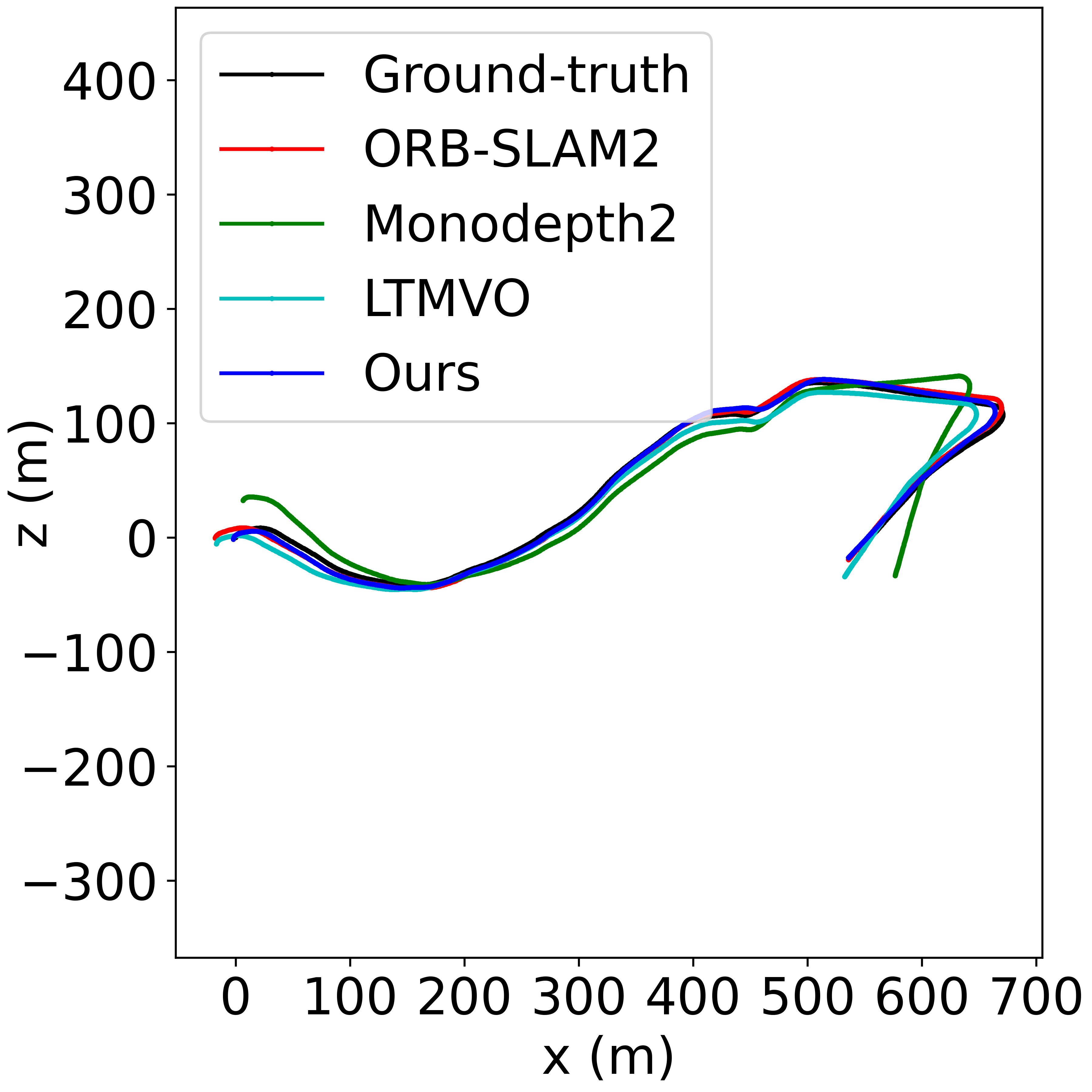} \\
        (a) Seq.09 &
        (b) Seq.10 \\
    \end{tabular}
    \caption{Qualitative evaluation on Sequence 09 and 10 of KITTI Odometry benchmark.}
    \label{fig_traj}
\end{figure}

In Fig. \ref{fig_traj}, we illustrate our global trajectories with other methods on Sequence 09 and 10. ORB-SLAM2 \cite{mur2017orb} shows large scale-drift errors on Sequence 09 while our method and LTMVO~\cite{zou2020learning} show superior global trajectories.
Our global RMSEs of trajectories are smaller than all other self-supervised methods, which indicates that the global consistency is well preserved in our method.

To show the relevance of results, we further report our results on Sequence 11-21 in Tab.~\ref{table_additions}.
Our method achieves competitive results among the learning-based methods.
In terms of the RMSE and $T_{rel}$ error, our method outperforms the other learning-based methods with a significant margin, and comparable even with ORB-SLAM2-M with loop closure.

{\tabcolsep=3pt
\begin{table}[t]
\caption{Average results on Sequence 11-21 of KITTI Odometry benchmark.}
\renewcommand\arraystretch{1.2}
\centering
\resizebox{1\linewidth}{!}
{
\begin{tabular}{ll|ccc}
\hline
    &  & RMSE & $T_{rel}$ & $R_{rel}$ \\[-2pt]
    & \multicolumn{1}{c|}{Method} & (m) & (\%) & (deg/100m) \\ \hline
\multicolumn{1}{l|}{\multirow{2}{*}{Geo.}}      & ORB-SLAM2-M (w/o LC) \cite{mur2017orb}        & 81.20          & 19.60          & 0.94          \\
\multicolumn{1}{l|}{}                           & ORB-SLAM2-M (w LC) \cite{mur2017orb}          & \textbf{44.09} & \textbf{12.96} & \textbf{0.71} \\ \hline
\multicolumn{1}{l|}{\multirow{4}{*}{Self Sup.}} & Monodepth2 \cite{godard2019digging}                  & 99.36          & 11.42          & 3.38          \\
\multicolumn{1}{l|}{}                           & SC-SfMLearner \cite{bian2019unsupervised}               & 156.66         & 19.04          & 5.77          \\
\multicolumn{1}{l|}{}                           & LTMVO \cite{zou2020learning}                      &  {\color[HTML]{3531FF} 73.18}          &  {\color[HTML]{3531FF} 7.08}           & {\color[HTML]{FE0000} \textbf{1.56}} \\
\multicolumn{1}{l|}{}                           & {\bf MLF-VO (Ours)}                        & {\color[HTML]{FE0000} \textbf{49.96}} & {\color[HTML]{FE0000} \textbf{6.44}}  &  {\color[HTML]{3531FF} 2.07}          \\ \hline
\end{tabular}
}
\label{table_additions}
\end{table}
}

{\tabcolsep=3pt
\begin{table}[t]
\caption{Pose inference time per image pair.}
\renewcommand\arraystretch{1.2}
\centering
\Huge
\resizebox{\linewidth}{!}
{
\begin{tabular}{l|c|c|c}
\hline
Method        & Resolution                    & Device      & Time (ms) \\ \hline
ORB-SLAM2-M \cite{mur2017orb}     & 376 $\times$ 1241 & 2-core CPU & 60        \\
LTMVO \cite{zou2020learning}         & 192 $\times$ 640 & GTX TitanXP & 70        \\
{\bf MLF-VO (Ours)}          & 192 $\times$ 640 & GTX 1080Ti  & \textbf{27}        \\ \hline
\end{tabular}
}
\label{table_runtime}
\end{table}
}

\para{Computation time analysis:}
In addition to the accuracy evaluation, we also compare the inference time of our method with other methods. We obtain the runtime of ORB-SLAM2 from their official report on KITTI Odometry benchmark, and the runtime of LTMVO from their public paper. Tab.~\ref{table_runtime} reports the comparison results with the image resolution and device. Given similar computational resources, our method is much faster than LTMVO~\cite{zou2020learning}.
We think this is because our network structure is simpler than LTMVO that employs an LSTM network.

\section{CONCLUSION}
\noindent
In this paper we have evaluated various strategies to fuse the RGB image and inferred depth image for ego-motion estimation. 
Through the validation on our baseline pipeline, we found several important observations for effectively fusing information from different modalities. 
We finally incorporate them to our pipeline and build a relative pose estimator that fuses modalities in multiple stages of the feature encoder and achieves state-of-the-art performance among self-supervised ego-motion estimation methods.

\para{Acknowledgement}. This work is partly supported by JSPS KAKENHI Grant Number 17H00744.

\bibliographystyle{IEEEtran}
\bibliography{shortstrings,reference}

\end{document}